 \documentclass[pmlr,twocolumn,10pt]{jmlr} 

\usepackage{booktabs}
\usepackage{siunitx}
\usepackage{amssymb}
\usepackage{pifont}
\usepackage{multirow}
\usepackage[switch]{lineno}

\usepackage{listings}
\usepackage{xcolor}
\usepackage{multicol}

\usepackage{longtable}

\lstdefinestyle{customstyle}{
    breaklines=true,
    breakatwhitespace=true,
    basicstyle=\ttfamily\footnotesize,
    commentstyle=\color{gray},
    keywordstyle=\color{blue},
    stringstyle=\color{red},
    numbers=left,
    numberstyle=\tiny\color{gray},
    numbersep=5pt,
    tabsize=2,
    extendedchars=true,
    backgroundcolor=\color{white},
    showspaces=false,
    showtabs=false,
    frame=single,
    rulecolor=\color{black},
    captionpos=b,
    breaklines=true,
    breakatwhitespace=false,
    escapeinside={\%*}{*)},
    keepspaces=true,
}

\newcommand{\equal}[1]{{\hypersetup{linkcolor=black}\thanks{#1}}}

\theorembodyfont{\upshape}
\theoremheaderfont{\scshape}
\theorempostheader{:}
\theoremsep{\newline}

 \title[Head CT Ontology Normalized Evaluation]{HeadCT-ONE: Enabling Granular and Controllable Automated Evaluation of Head CT Radiology Report Generation}

\author{%
\Name{Julián N. Acosta, MD}\equal{These authors contributed equally} \Email{julian\_acosta@hms.harvard.edu}\\
\addr Harvard University, USA
\AND
\Name{Xiaoman Zhang, PhD}\footnotemark[1] \Email{xiaoman\_zhang@hms.harvard.edu}\\
\addr Harvard University, USA
\AND
\Name{Siddhant Dogra, MD} \Email{siddhant.dogra@nyulangone.org}\\
\addr NYU Langone Health, USA
\AND
\Name{Hong-Yu Zhou, PhD} \Email{hongyu\_zhou@hms.harvard.edu}\\
\addr Harvard University, USA
\AND
\Name{Sam Payabvash, MD} \Email{sp4479@cumc.columbia.edu}\\
\addr Columbia University, USA
\AND
\Name{Guido J. Falcone, MD, ScD, MPH} \Email{guido.falcone@yale.edu}\\
\addr Yale University, USA
\AND
\Name{Eric K. Oermann, MD} \Email{eric.oermann@nyulangone.org}\\
\addr NYU Langone Health, USA
\AND
\Name{Pranav Rajpurkar, PhD} \Email{pranav\_rajpurkar@hms.harvard.edu}\\
\addr Harvard University, USA
}

\begin{document}

\maketitle

\begin{abstract}
We present Head CT Ontology Normalized Evaluation (HeadCT-ONE), a metric for evaluating head CT report generation through ontology-normalized entity and relation extraction. HeadCT-ONE enhances current information extraction derived metrics (such as RadGraph F1) by implementing entity normalization through domain-specific ontologies, addressing radiological language variability. HeadCT-ONE compares normalized entities and relations, allowing for controllable weighting of different entity types or specific entities. Through experiments on head CT reports from three health systems, we show that HeadCT-ONE's normalization and weighting approach improves the capture of semantically equivalent reports, better distinguishes between normal and abnormal reports, and aligns with radiologists' assessment of clinically significant errors, while offering flexibility to prioritize specific aspects of report content. Our results demonstrate how HeadCT-ONE enables more flexible, controllable, and granular automated evaluation of head CT reports.
\end{abstract}

\paragraph*{Code Availability}
The code is publicly available on GitHub.\footnote{\url{https://github.com/rajpurkarlab/HeadCT-ONE}}

\section{Introduction}
\label{sec:intro}

\begin{figure*}[t]
    \centering
    \includegraphics[width=1\linewidth]{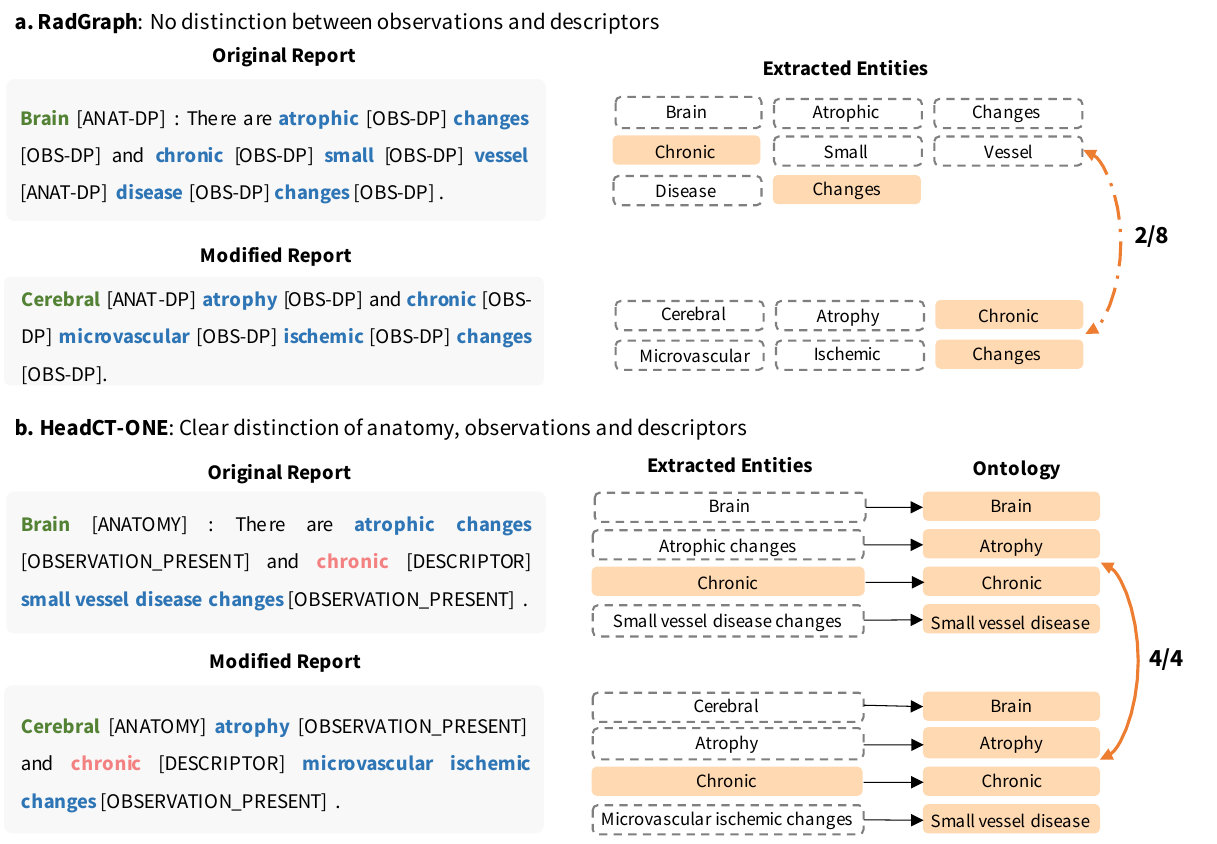}  \vspace{-10pt}
    \caption{Comparison of entity extraction results between RadGraph and HeadCT-ONE for two reports with different writing styles but identical findings. With ontology normalization, HeadCT-ONE's extractions are identical for both reports, while RadGraph only matches 1/4 of the entities.}
    \label{fig:teaser}
\end{figure*}

Automated radiology report generation, leveraging artificial intelligence (AI) to produce descriptive text from medical images, has gained significant attention due to its potential to streamline clinical workflows and improve patient care \citep{Moor2023-zo, Rajpurkar2023-ut, Sloan_2024}. As AI systems approach human-like performance in generating radiology reports from images, the development of robust, automated evaluation metrics becomes paramount. However, evaluating AI-generated radiology reports presents unique challenges due to the specialized nature of medical language and the semantic complexity of these reports.

Traditional natural language generation metrics, such as BLEU~\citep{papineni2002bleu} and ROUGE~\citep{lin2004rouge}, often fall short when applied to radiology, primarily focusing on lexical overlap and struggling to capture nuanced, semantic similarities crucial in medical reporting. One approach to address this limitation is the use of classification labels, such as the CheXpert F1 score \citep{irvin2019chexpertlargechestradiograph}, which evaluates the presence or absence of specific clinical findings. However, while this method sidesteps lexical variation issues, it fails to capture the level of detailed crucial in radiology reports.
More sophisticated approaches include RadGraph F1 \citep{Yu2023-xr}, an evaluation metric based on RadGraph \citep{jain2021radgraphextractingclinicalentities} that extracts entities and relations in reference and candidate reports and measure their overlap, providing a more meaningful evaluation but lacking robustness to variations in medical language (Figure~\ref{fig:teaser}), and embedding-based techniques \citep{zhang2020bertscoreevaluatingtextgeneration, pmlr-v158-endo21a}, which aim to capture semantic similarities by comparing vector representations of text, but output a single number, lacking explainability.

Large Language Model (LLM)-based evaluation methods have recently emerged as a promising approach \citep{huang2024fineradscoreradiologyreportlinebyline, chaves2024clinicallyaccessibleradiologyfoundation, xie2024doclensmultiaspectfinegrainedevaluation, ostmeier2024greengenerativeradiologyreport}, being more robust to stylistic differences \citep{banerjee2024rexamineglobalframeworkuncoveringinconsistencies} and aligning well with radiologists. However, current LLM-based metrics only ascertain error type and their clinical significance, limiting the potential for more granular evaluation, for example, by weighting on different pathologies, anatomical locations or descriptors, which may be more relevant for different clinical scenarios or distinct models being evaluated.

We argue that automated information extraction, coupled with accurate normalization to domain-specific ontologies can address these limitations by providing a standardized framework for medical terminology, enabling fine-grained categorization of concepts, and allowing for customizable weighting of different entities. Further, this approach could still leverage the natural language understanding of LLMs at different steps \citep{Gilbert2024-pv}, offering a more robust, explainable, and flexible approach to radiology report evaluation.

In this work: 

\begin{enumerate}
    \item We present a new information extraction framework that extends the RadGraph radiology report information extraction framework by introducing entity normalization, leveraging domain-specific ontologies for observations, anatomical terms, and descriptors. While this pipeline is focused on head CT, it is easily extendable to other CT modalities.

    \item We introduce HeadCT-ONE (Ontology normalized evaluation), a radiology report generation evaluation metric measuring overlap between these normalized entities and relations in candidate and reference reports, allowing the metric to be robust to differences in radiological language. We demonstrate that HeadCT-ONE captures similarities between radiology reports more effectively, showing increased robustness to variations in style and language due to the normalization process, facilitating the distinction between normal and abnormal reports.

    \item We introduce a weighting mechanism for different entity types or specific entities, allowing for the creation of metrics that focus on particular aspects of AI-generated reports. This controllability enables more nuanced evaluation tailored to specific clinical scenarios or model limitations. We illustrate how weighting on different entity types or specific entities allows our metric to focus on distinct issues in radiology report generation, and align better with radiologists' assessments of clinically significant errors.
\end{enumerate}

Our work represents a significant step towards more flexible and aligned automated evaluations of radiology report generation, facilitating the development and clinical evaluation of AI systems for head CT radiology report generation.

\section{Related Work}
There have been multiple efforts to develop reliable metrics for radiology report generation. Traditional natural language generation metrics, such as BLEU, ROUGE, and METEOR primarily focus on lexical overlap and struggle to capture the nuanced, semantic similarities that are crucial in medical reporting.
\citep{Yu2023-xr} introduced RadGraph F1, a more clinically meaningful evaluation metric based on extracting entities and relations from reports using RadGraph \citep{jain2021radgraphextractingclinicalentities}. However, the inherent variability in radiological language---where multiple phrases can express the same clinical finding---poses a challenge for this type of metric. This variability can lead to the underestimation of similarity between clinically equivalent but lexically different reports.

Embedding-based evaluation metrics, such as BERTScore \citep{zhang2020bertscoreevaluatingtextgeneration} and SemB score \citep{pmlr-v158-endo21a}, have attempted to address these limitations by leveraging pre-trained language models to capture semantic similarities. While these approaches offer improvements over traditional lexical overlap metrics, they still face challenges in the medical domain. These metrics often lack explainability, making it difficult for clinicians to interpret and trust the scores. Additionally, when applied at the report level, they struggle to provide granular insights or allow for fine-grained control over the evaluation process, which is crucial in the nuanced field of radiology.

Large Language Model (LLM)-based evaluation approaches have emerged as a promising alternative for assessing radiology reports, with multiple methods recently introduced, including FineRadScore \citep{huang2024fineradscoreradiologyreportlinebyline}, CheXprompt \citep{chaves2024clinicallyaccessibleradiologyfoundation}, DocLens \citep{xie2024doclensmultiaspectfinegrainedevaluation} and GREEN \citep{ostmeier2024greengenerativeradiologyreport}, offering several advantages over traditional metrics. These methods demonstrate improved alignment with human judgment and exhibit robustness to lexical variations, capturing semantic similarities that might elude conventional metrics. However, LLM-based evaluations come with their own set of challenges, being limited to the categorization of error types and their clinical significance, which they struggle to capture accurately. Further, running these metrics at scale can be prohibitively expensive, especially when dealing with large datasets common in medical imaging. Moreover, the use of LLMs raises significant privacy concerns, particularly when handling sensitive medical data or proprietary datasets.

\citet{zhao2024ratescoremetricradiologyreport} introduced RaTEScore, demonstrating promise in utilizing a named entity recognition and synonym disambiguation approach, outperforming other metrics on its alignment with radiologists' scores. However, RaTEScore does not leverage ontologies nor defines a descriptor entity type, which limit the potential controllability of this score.

\section{Method}
\subsection{Dataset}

\paragraph{Primary dataset.} We leverage a large proprietary dataset of 101,319 head CT and their corresponding radiology reports from three health systems in the United States and multiple sites within them, containing studies from 35,380 patients. Mean age at the time of study is 65.7 (SD 18.7), and 57\% of the studies are from female patients. This dataset is private and will not be available to other researchers. 
\paragraph{NER models training data.} We randomly sampled 2,000 radiology reports from this dataset to train our NER models.
\paragraph{Experiments data.}
Original reports. We manually selected 1 normal and 1 abnormal report per site for the 20 sites with most studies in the dataset. Additionally, we randomly sampled a non-overlapping group of 400 reports from these same 20 distinct sites.

Modified reports. We prompt GPT-4-o through the Azure OpenAI secure API to obtain 5 different versions of these reports: rephrased, any error, observation errors, anatomical errors, and descriptor errors. Prompts are shown in the Appendix.

\subsection{Ontology}
Our team developed a specialized ontology for the three main entity types of the expanded information extraction schema.
\paragraph{Observation ontology}: We built on the ontology tree published by \citet{Buchlak2023-yh}. Specifically, we simplified the ontology to remove terms related to anatomy (e.g., basal cistern effacement $\rightarrow$ effacement) or descriptors (e.g., acute hemorrhage $\rightarrow$ hemorrhage), which will be already captured as other entity types in our schema, and introduced terms not covered by this ontology tree. This observation ontology covers elemental findings present in head CT reports, including pathological findings, devices or surgical changes. The resulting ontology tree can be seen in the Appendix.
\paragraph{Descriptor ontology}: We developed a comprehensive and extendable ontology of features associated with radiological findings or anatomical structures in medical imaging. These features include common descriptors such as quantity, size, shape, severity, temporality (e.g., acute, chronic), among many other categories. The full ontology, which can be extended, is presented on the Appendix.
\paragraph{Anatomy ontology}: We utilize the Foundational Model of Anatomy (FMA) \citep{Rosse2003-mm}, an well-validated ontology of anatomical knowledge. We limit our pipeline to use entities reachable by starting with anatomical structures related to head CTs (i.e., Head, and related anatomical structures, such as Epidural Space), with a maximum depth of 5 to avoid entities too granular too appear in reports.

\subsection{Information Extraction Framework}
We present a new information extraction framework that extends RadGraph~\citep{jain2021radgraphextractingclinicalentities} by introducing a distinct "descriptor" entity type. This separates descriptors from observations, which were previously combined in RadGraph. Our framework retains RadGraph's anatomy and observation entities while adding this new category, enhancing the precision of information extraction from radiology reports. Our framework utilizes four relations: suggestive of, associated with, located at, and modify.
To develop models capable of extracting information from head CT reports according to our schema,
We use GPT-4~\citep{achiam2023gpt} to generate labeled entities and relations for a subset of our data. The prompts used for annotation are provided in the Appendix.
Based on the annotated data, we train a Named Entity Recognition (NER) model using the Princeton University Relation Extraction (PURE) architecture~\citep{zhong2020frustratingly}. This architecture employs a pipeline approach, decomposing the tasks of entity recognition and relation extraction into separate subtasks.
We apply the trained model as the first step of HeadCT-ONE to extract all entities and relations from the reports.

\subsection{Ontology Normalization}
Following information extraction, we map all extracted entities to our predefined ontologies to ensure consistency and enable standardized analysis.
For observation and anatomy entities, we leverage BioLORD-2023~\citep{remy2024biolord}, a state-of-the-art multilingual model, to generate high-fidelity embeddings of all ontology terms. We then match each extracted entity to the ontology term with the highest embedding similarity within its entity type. 
We also apply some minor rule-based postprocessing pipelines to refine the extracted entities. For example, ``frontoparietal'' would be separated into ``frontal'' and ``parietal'' to align with our ontology structure. 
For descriptors, given their limited categorical nature, we employ GPT-4 to classify extracted entities into our predefined descriptor ontology categories using a prompt provided in the Appendix. This normalization process facilitates consistent and controllable analyses across diverse reports.

\subsection{HeadCT-ONE Metric}
HeadCT-ONE metric assesses the model's ability to correctly identify entities (including the new descriptor type) and their relations in radiological reports.
The HeadCT-ONE metric is calculated as the average of two weighted F1 scores: one for entity recognition and one for relation extraction. 
Weights are assigned to individual items (entities or relations) based on criteria such as entity type, relation type, and clinical relevance.
This weighting mechanism allows the metric to be tailored to specific evaluation needs.

Define $e_{gt}$ and $e_{pred}$ as the sets of ground truth and predicted entities, and $r_{gt}$ and $r_{pred}$ be the sets of ground truth and predicted relations, respectively.
The metric is calculated as the average of two weighted F1 scores:
$$\text{HeadCT-ONE} = \frac{F1(e_{gt}, e_{pred}) + F1(r_{gt}, r_{pred})}{2},$$
where $F1$ is a weighted F1 score calculated based on precision ($P$) and recall ($R$). 

\subsection{Institutional Review Board (IRB)}
This research utilizes anonymized data and does not require IRB approval.

\section{Results}

\subsection{Analysis on Original Reports}

\begin{table}[!t]
\centering
\small
\setlength{\tabcolsep}{3pt}
\begin{tabular}{lccc}
\toprule
Metric & Weight & Normal & Normal-Abnormal \\
\midrule
GREEN & \ding{55} & 0.691 & 0.174 \\
RaTEScore & \ding{55} & 0.713 & 0.421  \\
RadGraph & \ding{55} & 0.328 & 0.160 \\
HeadCT-ONE & \ding{55} & 0.429 & 0.169 \\
RadGraph & \ding{51} & 0.224 & 0.149 \\
HeadCT-ONE & \ding{51} & \textbf{0.903} & \textbf{0.903} \\
\bottomrule
\end{tabular}
\caption{Comparison of average metric scores across 20 sites for (1) two normal reports and (2) the difference between normal pairs and normal-abnormal pairs. ``Weight'' indicates whether the metric uses weighting.}
\label{tab:normal}
\end{table}

\begin{figure}[!t]
    \centering
    \includegraphics[width=1\linewidth]{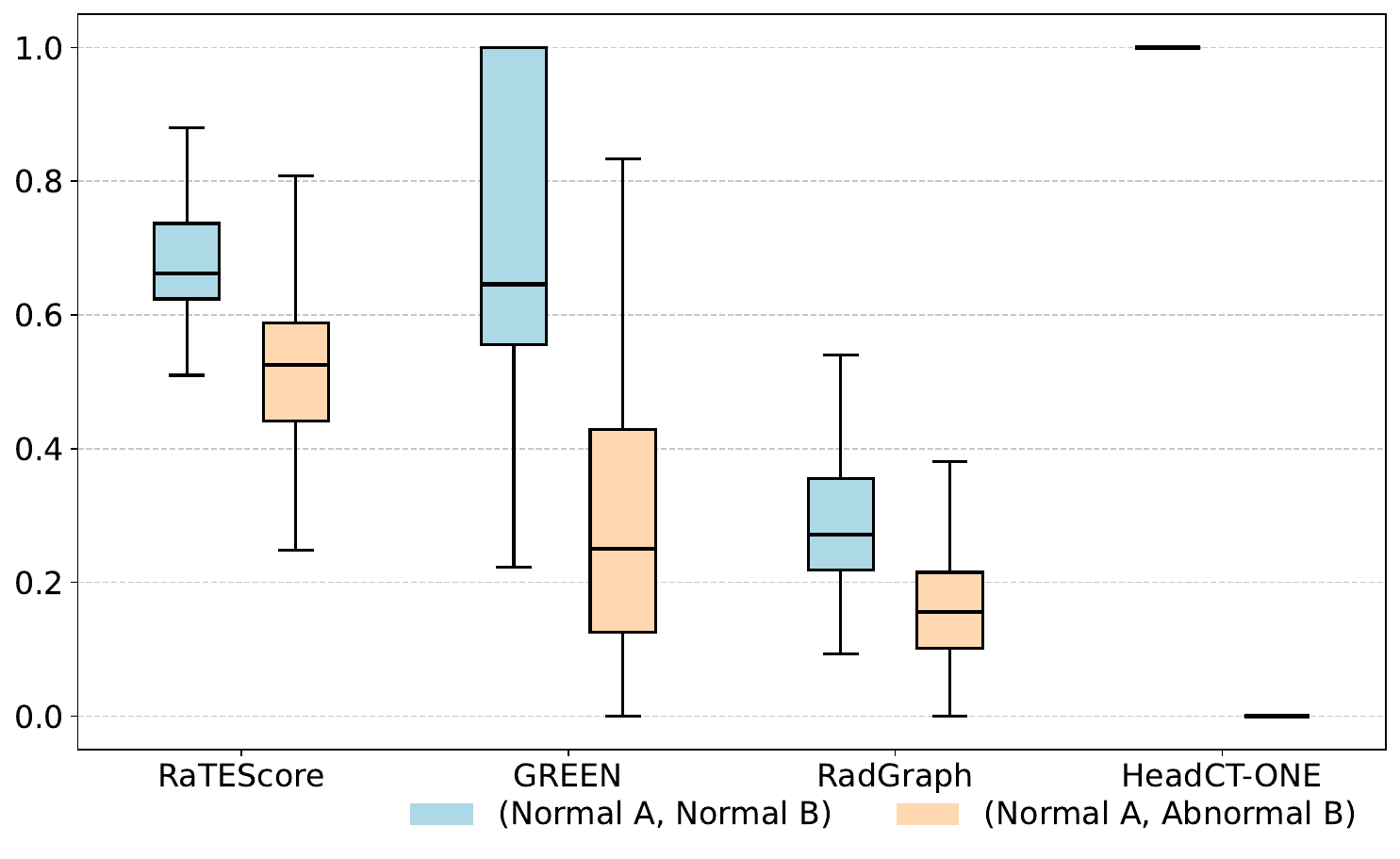}
    \caption{Scores of different report generation metrics when comparing two normal reports (light blue) or one normal and one abnormal report (light orange).}
    \label{fig:boxplot}
\end{figure}

\paragraph{Ontology normalization enables HeadCT-ONE to consistently identify similar concepts across diverse radiology reports.}
To evaluate this, we conducted a comparative analysis using reports from 20 different sites. 
For each site, one normal report and one abnormal report were selected. 
Our experiment tested how well various metrics could recognize similarities between normal reports and distinguish between normal and abnormal reports, regardless of differences in writing styles.

Table~\ref{tab:normal} presents the average scores for these comparisons. For each site, we computed a ``Normal'' score by comparing its normal report with normal reports from other sites, and a ``Normal-Abnormal" score by calculating the difference between normal-normal and normal-abnormal comparisons. This ``Normal-Abnormal" score effectively measures each metric's ability to differentiate between normal and abnormal reports while avoiding biases from consistently high or low scores. 
Additionally, Figure~\ref{fig:boxplot} presents a boxplot visualization of results across 20 sites, providing insight into score distribution. 

Our results demonstrate that HeadCT-ONE consistently outperformed RadGraph, achieving higher scores across all metrics. This superiority suggests that normalizing the data using ontologies significantly improves the overall analysis performance. 

\begin{table*}[!t]
\centering
\footnotesize
\setlength{\tabcolsep}{2pt}
\begin{tabular}{lccccccccccccc}
\toprule
\multirow{2}{*}{\textbf{Metric}} & \multicolumn{4}{c}{\textbf{Weight}} & \multicolumn{9}{c}{\textbf{F1 Score}} \\
\cmidrule(lr){2-5} \cmidrule(lr){6-14}
 & OBS-P & OBS-A & ANAT & DESC & Reph & Any & Obs & Ana  & Des & Reph-Any. & Reph-Obs & Reph-Ana & Reph-Des  \\
\midrule
\textbf{GREEN} & / & / & / & / & 0.820 & 0.639 & 0.707 & 0.703 & 0.679 & 0.182 & 0.114 & 0.118 & 0.142 \\
\textbf{RaTEScore} & / & / & / & / & 0.784 & 0.733 & 0.730 & 0.759 & 0.744 & 0.051 & 0.054 & 0.025 & 0.040 \\
\midrule
\textbf{RadGraph} & 1 & 1 & 1 & / & 0.426 & 0.341 & 0.349 & 0.343 & 0.354 & 0.085 & 0.077 & 0.083 & 0.072 \\
 & 1 & 0 & 0 & / & 0.393 & 0.334 & 0.328 & 0.380 & 0.307 & 0.059 & 0.065 & 0.013 & 0.086\\
 & 1 & 1 & 0 & / & 0.360 & 0.329 & 0.324 & 0.354 & 0.310 & 0.032 & 0.036 & 0.006 & 0.050\\
 & 0 & 0 & 1 & / & 0.516 & 0.448 & 0.467 & 0.419 & 0.497 & 0.068 & 0.050 & 0.097 & 0.019 \\
\midrule
\textbf{HeadCT-ONE} & 1 & 1 & 1 & 1 & 0.310 & 0.274 & 0.283 & 0.279 & 0.274 & 0.037 & 0.027 & 0.031 & 0.036 \\
\textbf{w/o Ontology} & 1 & 0 & 0 & 0 & 0.440 & 0.241 & 0.237 & 0.397 & 0.313 & 0.199 & 0.203 & 0.044 & 0.128 \\
 & 1 & 1 & 0 & 0 & 0.243 & 0.217 & 0.215 & 0.230 & 0.221 & 0.027 & 0.029 & 0.013 & 0.023 \\
 & 0 & 0 & 1 & 0 & 0.346 & 0.287 & 0.319 & 0.255 & 0.327 & 0.060 & 0.028 & 0.091 & 0.019 \\
 & 0 & 0 & 0 & 1 & 0.342 & 0.311 & 0.314 & 0.337 & 0.281 & 0.032 & 0.029 & 0.006 & 0.062 \\
\midrule
\textbf{HeadCT-ONE} & 1 & 1 & 1 & 1 & 0.631 & 0.558 & 0.571 & 0.581 & 0.571 & 0.073 & 0.060 & 0.051 & 0.060 \\
 & 1 & 0 & 0 & 0 & 0.719 & 0.477 & 0.468 & 0.657 & 0.570 & \textbf{0.242} & \textbf{0.251} & 0.062 & \textbf{0.149} \\
 & 1 & 1 & 0 & 0 & 0.671 & 0.597 & 0.597 & 0.645 & 0.625 & 0.074 & 0.074 & 0.026 & 0.046 \\
 & 0 & 0 & 1 & 0 & 0.544 & 0.446 & 0.495 & 0.407 & 0.512 & 0.099 & 0.050 & \textbf{0.138 }& 0.032 \\
 & 0 & 0 & 0 & 1 & 0.646 & 0.590 & 0.587 & 0.635 & 0.554 & 0.056 & 0.060 & 0.011 & 0.092 \\
\bottomrule
\end{tabular}
\caption{Comparison of F1 scores for various metrics and weighting schemes applied to 400 radiology reports, including original, rephrased, and error-inserted versions. Weighting schemes (OBS-P: Observation-present, OBS-A: Observation-absent, ANAT: Anatomy, DESC: Descriptor) use 1 or 0 to indicate the weighing number of related entities and relations. Columns show scores for rephrased reports (Reph) and reports with errors of any type (Any), with observation errors (Obs), anatomy errors (Ana) and descriptor errors (Des). The last four columns (Reph-Any, Reph-Obs, Reph-Ana, Reph-Des) represent score differences between (Original, Rephrased) and (Original, Error-Inserted Version). }
\label{tab:metric}
\end{table*}

\paragraph{Weighting on OBS-P enables HeadCT-ONE to easily distinguish between normal and abnormal reports.}
As illustrated in Table~\ref{tab:normal} and Figure~\ref{fig:boxplot}, after adding weights to all entities and relations related to observation-present, HeadCT-ONE demonstrated an improved ability to distinguish between normal and abnormal reports. This is evidenced by HeadCT-ONE achieving the highest score in the (Normal A, Normal B) - (Normal A, Abnormal B) comparison.
This indicates that the weighted approach in HeadCT-ONE is more sensitive to the differences between normal and abnormal reports.

The effectiveness of this weighting strategy is further demonstrated by HeadCT-ONE's performance on normal pairs, with a perfect score of 1 for normal pairs in 19 out of 20 sites. 
The one site where HeadCT-ONE did not achieve a perfect score was due to an error assignment of ``aeration'' as ``observation\_present'' in the sentence  ``The bone windows demonstrate normal aeration of the paranasal sinuses and mastoid air cells'', leading to this discrepancy. 
This example suggests that there is room for further refinement in its entity recognition accuracy, particularly for terms that may have nuanced meanings depending on their context.

\subsection{Analysis on Modified Reports}
\paragraph{The weighting mechanism for entity types allows HeadCT-ONE to focus on specific aspects of AI-generated reports.}
HeadCT-ONE's weighting mechanism for entity types provides a powerful tool for customizing the analysis of AI-generated radiology reports. As demonstrated in Table 2, this allows us to prioritize the detection of specific error types by adjusting the weights assigned to different entity categories.
For example, when there is a particular concern about observation-related errors, increasing the weight of the observation-present (OBS-P) category enhances HeadCT-ONE's sensitivity to these types of discrepancies. This is evident in the results where the weighting scheme (1, 0, 0, 0) for (OBS-P, OBS-A, ANAT, DESC) yields the highest F1 score differences for observation errors (Reph-Obs: 0.251) and general errors (Reph-Any: 0.242). 
Similarly, if the focus is on detecting anatomy-related errors, assigning a higher weight to the anatomy (ANAT) category improves performance in this area. The weighting scheme (0, 0, 1, 0) results in the highest F1 score difference for anatomy errors (Reph-Ana: 0.149), demonstrating HeadCT-ONE's ability to adapt to this specific focus.

\begin{figure*}[t]
    \centering
    \includegraphics[width=1\linewidth]{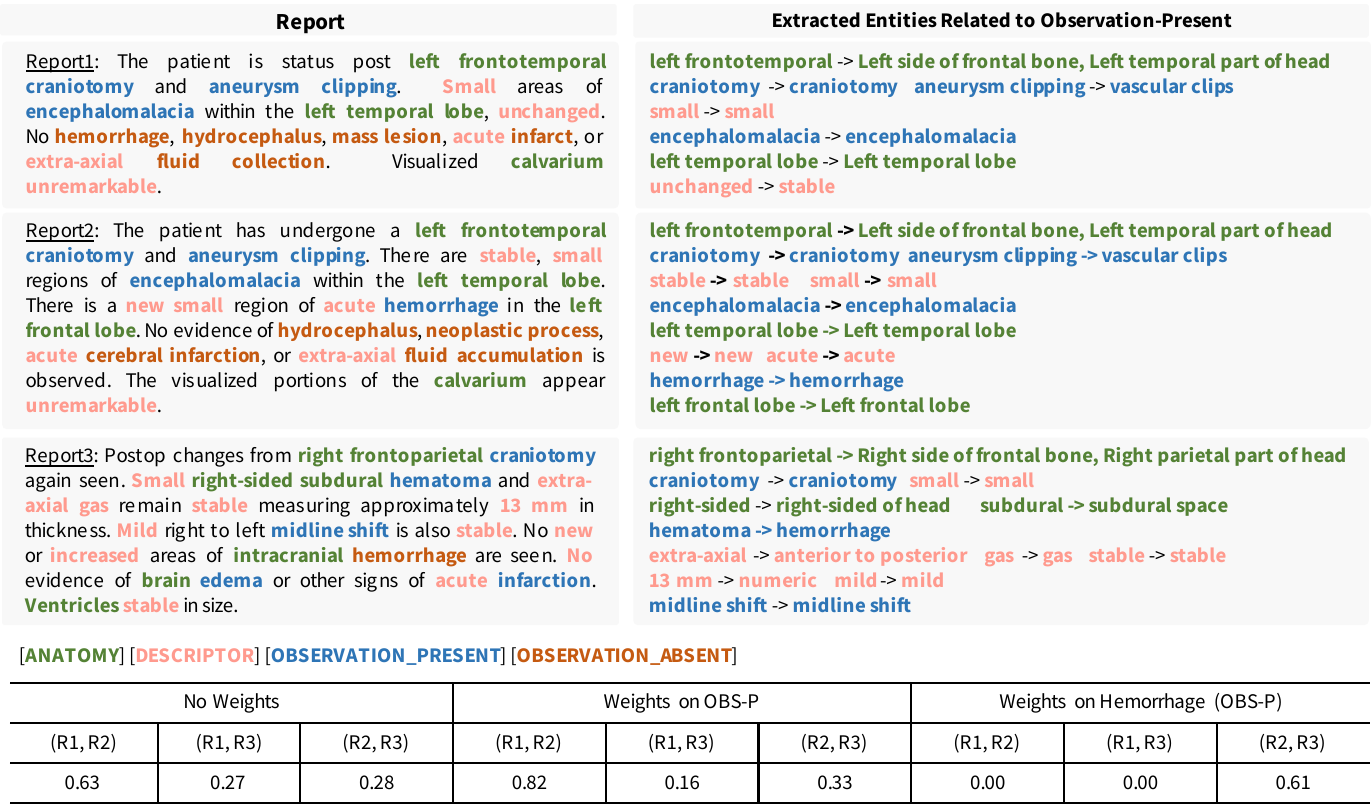}
    \caption{Case study. F1 scores between report pairs (R1,R2), (R1,R3), and (R2,R3) under different weighting schemes. No weights shows baseline similarities. Weights on OBS-P (observation-present) increases similarity for reports sharing many positive observations. Weights on 'Hemorrhage' (OBS-P) highlights specific differences in hemorrhage mentions, drastically changing similarities between otherwise similar reports.}
    \label{fig:casestudy}
\end{figure*}

\paragraph{The weighting of specific entities enables precise identification of targeted error types in radiology reports, enhancing its controllability.}
The application of entity-specific weights in HeadCT-ONE demonstrates the ability to focus on particular aspects of radiology reports, thereby improving error detection in targeted areas.
As shown in Figure \ref{fig:casestudy}, when applying weights to any observation-present entities, two reports sharing many positive observations except for hemorrhage (R1, R2) score high (0.82). However, this score drops to 0 when weighting only on specific `hemorrhage' observation-present entities, highlighting their key difference. Conversely, a different pair of reports with more distinct observations but both mentioning different types of hemorrhage (R2, R3) scores lower (0.33) when weighting on any observation-present entities. This pair's score increases to 0.61 when weighting specifically on hemorrhage, though it doesn't reach 1.0 due to the different types of hemorrhage mentioned. 

\begin{table}[!t]
\centering
\small
\setlength{\tabcolsep}{5pt}
\begin{tabular}{llcccc}
\toprule
Metric & Weight & w/o Sig. & w Sig. & Diff. \\
\midrule
GREEN & \ding{55} & 0.743 & 0.688 & 0.055** \\
RaTEScore & \ding{55} & 0.750 & 0.744 & 0.006* \\
RadGraph & \ding{55} & 0.366 & 0.371 & -0.005* \\
HeadCT-ONE & OBS-P & 0.716 & 0.543 & 0.173 \\
HeadCT-ONE & Top 5 & 0.732 & 0.497 & 0.235 \\
HeadCT-ONE & Top 10 & 0.706 & 0.409 & \textbf{0.297} \\
\bottomrule
\end{tabular}
\caption{Scores for metrics on reports with and without significant errors (Sig.). `Diff.' shows the difference between `w/o Sig.' and `w Sig.'. `Weight' indicates whether the metric uses weighting. * p $<$ 0.05, ** p $<$ 0.01 compared to HeadCT-ONE (Top 10). }
\label{tab:correlation}
\end{table}

\subsection{Alignment with Radiologists}
We investigate the correlation between automated metrics and radiologist assessments of radiology reports to validate the effectiveness of the HeadCT-ONE metric. 
A senior radiology resident classified modified reports as with and without significant errors compared to candidate reports.
We then calculated the scores for various metrics on modified reports vs. candidate reports,  The results are presented in Table~\ref{tab:correlation}.
We explore HeadCT-ONE with weighting only on observation-present entities (``OBS-P'') and applying higher weight to the top 5 or top 10 most frequently negated observation-present entities (``Top 5'' and ``Top 10'' respectively).
The results demonstrate that HeadCT-ONE exhibits promising performance in distinguishing between reports with and without significant errors. 
Notably, the ``Top 10'' weighting strategy for HeadCT-ONE shows the largest difference (0.297) between scores for reports without significant errors (0.706) and those with significant errors (0.409). 
This analysis validates HeadCT-ONE's capacity to capture clinically relevant aspects of report quality and highlights the potential for adjusting the metric to better align with radiologist judgments.

\section{Discussion}
Our study introduces HeadCT-ONE, an ontology-enhanced information extraction-derived metric for head CT radiology report generation. Our results showcase that HeadCT-ONE offers significant improvements over existing metrics in terms of robustness, granularity, and controllability.

  \paragraph{Entity normalization enhances information extraction-derived metrics.}Medical ontologies have a widespread use in healthcare, but their integration into radiology report generation and evaluation is challenging. Widely used ontologies such as SNOMED-CT \citep{Gaudet-Blavignac2021-iu}, International Classification of Diseases (ICD) \citep{Who2005-wq} and Unified Medical Language System (UMLS) \citep{Bodenreider2004-he} do not generalize well to the language utilized in radiology reports. While RadLex \citep{Chepelev2023-cx} aims to address radiology-specific terminology, its scope may not fully encompass all subspecialty terms \citep{Datta2020-pz}, and its granularity can be inconsistent, sometimes lacking detail for certain findings while being overly specific for others, potentially affecting standardized usage across different radiologists and institutions. In our study, we modified an ontology tree specifically designed for head CT findings, and developed an ontology for radiology descriptors based on head CT but easily extendable to other CT study types. In addition, we use the FMA for anatomical ontology as we found it to match radiological terms better than other ontologies. In our experiments, we show that the normalization step improves the capability of HeadCT-ONE to identify similar reports regardless of variations of radiological language.

\paragraph{Entity weighting allows for controllable evaluation and better alignment with experts.}Radiology reports encompass a complex hierarchy of information, where the clinical significance of elements varies across different study types and analytical objectives. For instance, when focusing on diagnostic accuracy, the presence or absence of critical observations may be paramount, while the comprehensive use of descriptors would be relevant when evaluating report completeness. This variability extends to specific clinical scenarios: in a head trauma context, the evaluation might prioritize acute findings such as intracranial hemorrhage, midline shift, or skull fractures. Conversely, in the longitudinal follow-up of hydrocephalus, the focus may shift to changes in ventricular size and shunt positioning. This multifaceted variability in the importance of report elements poses a significant challenge for traditional evaluation metrics, which often treat all terms with equal weight. Our pipeline introduces a weighting mechanism that allows for focused evaluation on broad or specific entity types. This controllability enables more nuanced and clinically relevant assessments, as evidenced by the improved correlation with radiologist evaluations. Moreover, in recognizing the current limitations of generative AI in radiology, our approach allows for strategic leniency in certain areas. For instance, one could modulate the impact of discrepancies in quantitative descriptors like precise measurements, which AI systems may struggle to consistently reproduce. 

\paragraph{Limitations.}Our study is not without limitations. The normalization process, while effective, may occasionally misclassify terms. Further, our analyses were done on synthetic reports, which may not completely represent outputs from real head CT report generation models. Additionally, the creation of comprehensive ontologies for different radiological domains remains a challenging and time-consuming task requiring domain knowledge. Looking ahead, promising directions for future research include combining our information extraction and normalization approach with the increasing capabilities of LLMs. LLM have shown to perform well at information extraction \citep{liu-etal-2023-exploring-boundaries}. For normalization, LLMs could provide more sophisticated and accurate term classification, enhancing the overall reliability of HeadCT-ONE. Additionally, leveraging LLMs for data-driven ontology creation could significantly improve scalability, reducing the need for extensive domain knowledge and expert involvement. Furthermore, as accurate head CT report generation models become available, validating HeadCT-ONE on their outputs will be crucial to ensure its effectiveness in evaluating real-world generated reports.

In conclusion, HeadCT-ONE enhances AI-generated head CT report evaluation, offering improved robustness, granularity, and controllability, being easily extendable to other imaging modalities.


\bibliography{jmlr-sample}

\newpage

\appendix
\setcounter{table}{0} 
\renewcommand{\thetable}{A\arabic{table}} 

\onecolumn
\section{Defined ontologies}\label{apd:onto}
\begin{longtable}{>{\raggedright\arraybackslash}p{3cm}>{\raggedright\arraybackslash}p{2cm}>{\raggedright\arraybackslash}p{2cm}>{\raggedright\arraybackslash}p{8cm}}
\caption{Descriptors ontology categorization, showing examples for each category. The first level represents the main category, followed by a more specific second and third level where applicable.} \label{tab:descriptors_ontology} \\

\toprule
\textbf{First level} & \textbf{Second level} & \textbf{Third level} & \textbf{Examples} \\
\midrule
\endfirsthead

\multicolumn{4}{c}%
{\tablename\ \thetable{} -- continued from previous page} \\
\toprule
\textbf{First level} & \textbf{Second level} & \textbf{Third level} & \textbf{Examples} \\
\midrule
\endhead

\midrule \multicolumn{4}{r}{Continued on next page} \\
\endfoot

\bottomrule
\endlastfoot

quantity & numeric &  & 5 lesions,3 fractures \\
 & qualitative & single & isolated,solitary,single \\
 &  & multiple & several,multiple,numerous,a few \\
\midrule
size & numeric &  & 5 mm,10 cm \\
 & qualitative & very\_small & tiny,microscopic \\
 &  & small & small \\
 &  & medium & moderate,average,medium size,normal size \\
 &  & large & large,big,enlarged \\
 &  & very\_large & enormous,huge,very large,gigantic,very enlarged \\
\midrule
shape & regular & spherical & round,oval,ovoid \\
 &  & saccular & saccular \\
 &  & curvilinear & curvilinear \\
 &  & crescentic & crescentic \\
 &  & biconvex & biconvex \\
 &  & laminar & laminar,sheet-like,layer \\
 &  & tubular & tubular,cylindrical \\
 &  & fusiform & fusiform \\
 & irregular & lobulated & lobulated \\
 &  & spiculated & spiculated \\
 &  & amorphous & amorphous \\
\midrule
homogeneity & homogeneous &  & homogeneous \\
 & heterogeneous &  & heterogeneous \\
\midrule
density & hypodense &  & hypodense,hypoattenuation,hypodensity \\
 & isodense &  & isodense \\
 & hyperdense &  & hyperdense,hyperattenuation,hyperdensity \\
 & mixed &  & mixed density \\
\midrule
margin & well\_defined &  & circumscribed,well defined,well circumscribed,well delimited \\
 & poorly\_defined &  & ill-defined,poorly circumscribed \\
\midrule
severity & minimal &  & minimal \\
 & mild &  & mild \\
 & moderate &  & moderate \\
 & severe &  & severe \\
\midrule
temporality & acute &  & acute,new \\
 & subacute &  & subacute \\
 & chronic &  & chronic,old,remote \\
 & acute\_on\_chronic &  & acute on chronic \\
 & age\_indeterminate &  & age-indeterminate,unknown age \\
\midrule
distribution & localized &  & focal,localized \\
 & diffuse &  & diffuse \\
 & confluent &  & confluent \\
 & scattered &  & scattered \\
 & petechial &  & petechial \\
 & multifocal &  & multifocal \\
\midrule
enhancement & present & homogeneous & homogeneous enhancement \\
 & & heterogeneous & heterogeneous enhancement \\
 & & peripheral & peripheral enhancement \\
 & & central & central enhancement \\
 & & rim & rim-like enhancement \\
 & & patchy &  patchy enhancement \\
 & absent & & no enhancement \\
\midrule
certainty & definitely\_present &  & there is, there are,with \\
 & probably\_present &  & probably,likely \\
 & possibly\_present &  & possibly \\
 & uncertain &  & cannot rule out \\
 & definitely\_absent &  & no evidence of,there is no,without \\
\midrule
composition & gas &  & gas,gaseous,air \\
 & fluid & simple\_fluid & fluid-like,simple fluid \\
 &  & csf & csf \\
 &  & serous & serous \\
 &  & hemorrhagic & hemorrhagic \\
 &  & mucinous & mucinous,colloid \\
 & solid & soft\_tissue & soft-tissue density \\
 &  & fatty & fatty \\
 &  & fibrous & fibrous \\
 &  & calcified & calcific density,calcified \\
 &  & sclerotic & sclerotic \\
 & mixed &  & mixed,semisolid,fluid and solid components,solid with hemorrhagic components \\
\midrule
complexity & simple &  & simple \\
 & complex &  & complex \\
\midrule
change & resolution &  & resolution,resolved,cleared,disappeared \\
 & improvement &  & improved,improving \\
 & increase &  & increased,increase in size,larger,increasing \\
 & decrease &  & decreased,decrease in size,decreasing,smaller \\
 & worsening &  & worsened,worsening \\
 & appearance &  & new,appeared,is now present \\
 & mixed\_change &  & one metastasis increased in size and the other one resolved \\
 & stable &  & stable,unchanged,similar,similar in appearance \\
\midrule
normalcy & normal &  & normal \\
 & abnormal &  & abnormal \\
\midrule
caliber & dilated &  & dilated ventricles,dilated vessels \\
 & normal &  & average lumen,normal cavity,normal calibre \\
 & reduced &  & narrowed artery,collapsed veins,collapsed ventricles \\
\midrule
malignancy\_status & definitely\_benign &  & benign \\
 & probably\_benign &  & probably benign lesion \\
 & indeterminate &  & indeterminate hypodensity \\
 & probably\_malignant &  & probably malignant,suspicious \\
 & definitely\_malignant &  & cancerous,malignant \\
\midrule
patency & patent &  & patent,clear \\
 & mostly\_patent &  & mostly patent \\
 & obstructed &  & obstruction,obstructed \\
 & occluded &  & occlusion \\
\midrule
occupancy & empty &  & empty,clear,well-aereated \\
 & partially\_filled &  & partially filled with.. \\
 & fully\_filled &  & full,fully filled \\
 & engorged &  & engorged \\
\midrule
integrity & intact &  & intact,unruptured \\
 & partially\_compromised &  & partially ruptured, partially disrupted \\
 & compromised &  & ruptured,disrupted,disruption \\
\midrule
direction & left\_to\_right &  & left-to-right \\
 & right\_to\_left &  & right-to-left \\
 & anterior\_to\_posterior &  & anterior-to-posterior \\
 & posterior\_to\_anterior &  & posterior-to-anterior \\
 & upwards &  & upwards \\
 & downwards &  & downwards \\
\midrule
component\_involved & mucosal &  & mucosal \\
 & muscular &  & muscular \\
 & osseous &  & osseous,bony \\
\midrule
position & normal\_position &  & normal position \\
 & abnormal\_position &  & displaced,abnormal position \\
\end{longtable}

\clearpage

\begin{longtable}{>{\raggedright\arraybackslash}p{5cm}>{\raggedright\arraybackslash}p{10cm}}
\caption{Findings ontology list, with synonyms when relevant.} \label{tab:finding_ontology} \\

\toprule
\textbf{Finding} & \textbf{Synonyms} \\
\midrule
\endfirsthead

\multicolumn{2}{c}%
{\tablename\ \thetable{} -- continued from previous page} \\
\toprule
\textbf{Finding} & \textbf{Synonyms} \\
\midrule
\endhead

\midrule \multicolumn{2}{r}{Continued on next page} \\
\endfoot

\bottomrule
\endlastfoot

infarct & ischemic stroke,infarction \\
hemorrhage & hematoma,bleed,blood \\
agenesis &  \\
lesion & mass,tumor,tumour \\
thickening &  \\
aneurysm &  \\
coils &  \\
subluxation &  \\
dissociation &  \\
effacement &  \\
calcification &  \\
thrombosis & clot,thrombus \\
beam\_hardening\_artefact &  \\
atrophy & involution,atrophic changes \\
cavum\_septum\_pellucidum &  \\
chiari\_1 &  \\
chiari\_2 &  \\
cochlear\_implant &  \\
cyst &  \\
colpocephaly &  \\
hydrocephalus &  \\
hypodensity &  \\
necrosis &  \\
craniotomy &  \\
collection &  \\
dbs\_electrodes &  \\
thinning &  \\
demyelination &  \\
diffuse\_axonal\_injury &  \\
venous\_gas &  \\
arachnoidocele &  \\
encephalitis &  \\
encephalomalacia &  \\
entrapment &  \\
external\_ventricular\_drainage & ventriculostomy catheter,ventriculostomy \\
exophthalmos &  \\
empyema &  \\
herniation &  \\
fracture &  \\
erosion &  \\
fibrous\_dysplasia &  \\
foreign\_body &  \\
fungal\_sinusitis &  \\
shape\_abnormality &  \\
heterotopia &  \\
hyperdense\_artery &  \\
hyperostosis &  \\
hypopneumatisation &  \\
hypoxic\_ischaemic\_encephalopathy &  \\
intracranial\_pressure\_monitor & icp \\
insular\_ribbon\_sign &  \\
silicone &  \\
debris &  \\
opacity &  \\
post\_surgical\_change &  \\
meningioma &  \\
metallic\_artefact &  \\
midline\_shift &  \\
movement\_artefact &  \\
mucocoele &  \\
non\_hemorrhagic\_contusion &  \\
optic\_neuritis &  \\
abscess &  \\
fat\_stranding &  \\
prosthesis &  \\
osteoma &  \\
otosclerosis &  \\
papilloedema &  \\
perivascular\_spaces &  \\
pseudo\_sah &  \\
resection\_cavity &  \\
schizencephaly &  \\
haemangioma &  \\
small\_vessel\_disease & white matter change,white matter changes,ischemic change,ischemic changes,microvascular changes,microvascular disease,microvascular change \\
stapes\_implants &  \\
ectopic\_air & emphysema,pneumocephalus \\
arthritis &  \\
dislocation &  \\
edema &  \\
transphenoidal\_surgery &  \\
vascular\_clips & aneurysm clips \\
vascular\_stents & stent,stents \\
venous\_infarct &  \\
venous\_thrombosis & cvt,venous sinus thrombosis,cerebral venous thrombosis \\
ventriculoperitoneal\_shunt & vp shunt \\
mass\_effect &  \\
loss\_of\_gray\_white\_matter\_differentiation &  \\
abnormality & pathology,finding,process,abnormalities \\
\end{longtable}

\clearpage
\section{Prompt for rephrasing and error introduction}\label{apd:ner}

\begin{lstlisting}[style=customstyle]
You are an expert radiologist tasked with rephrasing a given radiology report. Your objective is to rewrite the report using alternative medical terminology and sentence structures while maintaining the accuracy of the medical content. The rewritten report should sound distinctly different from the original but convey the same clinical information in a professional radiologist's style.

Please adhere to the following guidelines:
1. Utilize synonymous medical terms and phrases where appropriate.
2. Restructure sentences to present information in a different order or format.
3. Ensure that all key medical findings and diagnoses are preserved in the rewritten version.
4. Adapt the level of detail to match the original report.

Here's an example to illustrate the task:

Original Report:
FINDINGS: The lungs are clear without focal consolidation. No pleural effusion or pneumothorax is seen. The heart size is normal. The mediastinum is unremarkable.

Rewritten Report:
FINDINGS: Pulmonary fields demonstrate no evidence of focal opacities or airspace disease. Pleural spaces are free of fluid collections or air. Cardiac silhouette is within normal limits. Mediastinal structures appear unremarkable.

Please provide the rewritten report, maintaining medical accuracy while changing the wording and phrasing.
    
Original Radiology Report:
{report}
Rewritten Report:
\end{lstlisting}

\begin{lstlisting}[style=customstyle]
You are an expert radiologist tasked with modifying a given radiology report by introducing 1-3 errors while maintaining overall medical coherence. Your objective is to rewrite the report with subtle yet significant changes that could impact the interpretation of the findings.

Definitions:
- observation: A specific finding or abnormality noted in a radiology report. An observation typically describes a particular condition, anomaly, or feature identified during the imaging study.
- anatomical location: The specific area or structure in the body an observation is found, or an anatomical structure being described.
- descriptor: A characteristic or attribute used to describe an observation, such as size, shape, density, or severity.

Please adhere to the following guidelines:
1. Introduce 1-3 errors from any of the following categories:
   a) Observation errors:
      - Missing an abnormal observation: Omit an abnormal finding that was present in the original report.
      - Added an abnormal observation: Include a new abnormal finding that was not present in the original report.
      - Negated an abnormal observation: Change an abnormal finding to a negative or normal statement.
   b) Anatomical errors:
      - Incorrect anatomical location: Change the location of an observation to a different but plausible anatomical site, lobe, or region.
      - Incorrect anatomical side: Switch the side (left/right) of an observation.
   c) Descriptor errors:
      - Change one or more aspects of an observation, such as quantity, size, shape, homogeneity, density, margin, severity, temporality, distribution, enhancement, temporal change, certainty, composition, complexity, normalcy, caliber, malignancy status, patency, occupancy, integrity, direction, component involved, or position.
2. Ensure the errors are plausible and maintain medical coherence within the context of the report. You can modify the rest of the report to be consistent with the introduced errors.
3. Preserve the overall structure and style of the original report.

Please provide the modified report with introduced errors while maintaining overall medical coherence. Additionally, include information about which errors were introduced.

Your response should be in JSON format with two main keys: "modified_report" for the rewritten report, and "errors" for describing the introduced errors.

Original Radiology Report:
{report}

Please provide the rewritten report with 1-3 errors (observation, anatomical, or descriptor) in JSON format. Use the following structure:

{
    "modified_report": "Your rewritten report here",
    "errors": [
        {
            "type": "error type: observation (missing/added/negated), anatomical (incorrect location/incorrect side), or descriptor (specify type)",
            "description": "brief description of the error"
        },
        ...
    ]
}
\end{lstlisting}

\begin{lstlisting}[style=customstyle]
You are an expert radiologist tasked with modifying a given radiology report by introducing 1-3 observation errors while maintaining overall medical coherence. 
Your objective is to rewrite the report with subtle yet significant changes that could impact the interpretation of the findings.

Definitions:
- observation: A specific finding or abnormality noted in a radiology report. An observation typically describes a particular condition, anomaly, or feature identified during the imaging study.
- negation or normal statement: an statement commenting on the absence of a particular abnormal finding, or the normalcy of a structure.

Please adhere to the following guidelines:
1. Introduce 1-3 observation errors, which can be one of the following types:
   a) Missing an abnormal observation: Omit an abnormal finding that was present in the original report. DO NOT remove negative or normal statements.
   b) Added an abnormal observation: Include a new abnormal finding that was not present in the original report. DO NOT add negative or normal statements.
   c) Negated an abnormal observation: Change a abnormal finding to a negative or normal statement using the appropriate general negation or normal statements.
2. Avoid making changes to the anatomical location of observations (anatomical errors) and descriptors of these observations (e.g., severity).
3. Ensure the errors are plausible and maintain medical coherence within the context of the report. You can modify the rest of the report to be consistent with that error. The report should still make sense medically, even with the introduced errors.
4. Preserve the overall structure and style of the original report, even in the errors introduced.

Please provide the modified report with introduced observation errors while maintaining overall medical coherence. Additionally, include information about which errors were introduced.

Your response should be in JSON format with two main keys: "modified_report" for the rewritten report, and "errors" for describing the introduced errors.

Original Radiology Report:
{report}

Please provide the rewritten report with 1-3 observation errors in JSON format. Use the following structure:

{
    "modified_report": "Your rewritten report here",
    "errors": [
        {
            "type": "error type: missing/added/replaced/negated",
            "description": "brief description of the error"
        },
        ...
    ]
}
\end{lstlisting}

\begin{lstlisting}[style=customstyle]
You are an expert radiologist tasked with modifying a given radiology report by introducing 1-3 anatomical errors while maintaining overall medical coherence. 
Your objective is to rewrite the report with subtle yet significant changes that could impact the interpretation of the findings.

Definitions:
- anatomical error: An error in the description of the anatomical location of structures or findings in a radiology report.
- observation: A specific finding or abnormality noted in a radiology report. An observation typically describes a particular condition, anomaly, or feature identified during the imaging study.

Please adhere to the following guidelines:
1. Introduce 1-3 anatomical errors, which can be one of the following types:
   a) Incorrect anatomical location: Change the location of an observation to a different but plausible anatomical site, lobe or region for this particular observation, and study type.
   b) Incorrect anatomical side: Switch the side (left/right) of an observation.
2. Avoid making changes to the observations themselves (observation errors) and descriptors of these observations (e.g., severity).
3. Ensure the errors are plausible and maintain medical coherence within the context of the report. You can modify the rest of the report to be consistent with that error. The report should still make sense medically, even with the introduced errors.
4. Preserve the overall structure and style of the original report, even in the errors introduced.

Please provide the modified report with introduced anatomical errors while maintaining overall medical coherence. Additionally, include information about which errors were introduced.

Your response should be in JSON format with two main keys: "modified_report" for the rewritten report, and "errors" for describing the introduced errors.

Original Radiology Report:
{report}

Please provide the rewritten report with 1-3 anatomical errors in JSON format. Use the following structure:

{
    "modified_report": "Your rewritten report here",
    "errors": [
        {
            "type": "error type: incorrect location/incorrect side",
            "description": "brief description of the error"
        },
        ...
    ]
}
\end{lstlisting}

\begin{lstlisting}[style=customstyle]
You are an expert radiologist tasked with modifying a given radiology report by introducing 1-3 descriptor errors while maintaining overall medical coherence. 
Your objective is to rewrite the report with subtle yet significant changes that could impact the interpretation of the findings.

Definitions:
- descriptor error: An error in the description or characterization of an observation, finding or anatomical structure in a radiology report, without changing the observation itself or its anatomical location.
- observation: A specific finding or abnormality noted in a radiology report. An observation typically describes a particular condition, anomaly, or feature identified during the imaging study.

Please adhere to the following guidelines:
1. Introduce 1-3 descriptor errors, which can involve changing one or more of the following aspects of an observation:
   quantity, size, shape, homogeneity, density, margin, severity, temporality, distribution, enhancement, temporal change, certainty, composition, complexity, normalcy, caliber, malignancy status, patency, occupancy, integrity, direction, component involved, or position.
2. Avoid making changes to the observations themselves, adding observations (observation errors), or changing their anatomical locations (anatomical errors).
3. Ensure the errors are plausible and maintain medical coherence within the context of the report. You can modify the rest of the report to be consistent with that error. The report should still make sense medically, even with the introduced errors.
4. Preserve the overall structure and style of the original report, even in the errors introduced.

Please provide the modified report with introduced descriptor errors while maintaining overall medical coherence. Additionally, include information about which errors were introduced.

Your response should be in JSON format with two main keys: "modified_report" for the rewritten report, and "errors" for describing the introduced errors.
Original Radiology Report:
{report}

Please provide the rewritten report with 1-3 descriptor errors in JSON format. Use the following structure:

{
    "modified_report": "Your rewritten report here",
    "errors": [
        {
            "type": "error type: [descriptor type]",
            "description": "brief description of the error"
        },
        ...
    ]
}
\end{lstlisting}

\newpage
\section{Prompt for NER Annotation}\label{apd:ner}

\begin{lstlisting}[style=customstyle]
You are a radiologist performing clinical term extraction from the FINDINGS and IMPRESSION sections in the given radiology report. 
Here a clinical term can be in [`anatomy',`observation_present',`observation_absent',`device_present',`device_absent',`procedure',`descriptor']. 
`anatomy' refers to the anatomical body, such as `left frontal scalp', `paranasal sinus'; 
`observation_present' refers to findings, diseases are present according to the sentence, such as `haemorrhage', `lesion'; 
`observation_absent' refers to findings, diseases or medical devices are not present according to the sentence; 
`device_present' refers to medical devices are present according to the sentence, such as `ventricular drain', `stent', `clip'; 
`device_absent' refers to medical devices are not present according to the sentence; 
`procedure' refers to procedures are used to diagnose, measure, monitor or treat problems; 
`descriptor' refers to modifiers used to describe the observation, including categories as quantity (`single', `isolated', `multiple',etc), size (`large',`big',etc), shape (`lobulated',`round',etc), homogeneity(`homogeneous',`heterogeneous',etc), density (`isodense',`hypodense',etc), margin (`well-defined',`poor-defined',etc), severity (`mild',`moderate',`severe', etc), temporality (`acute',`new',`old',`age-indeterminate',etc), destribution (`confluent',`diffuse',etc), enhancement, change (`decreased',`worsened',etc), certainty (`probably',`like'), composition (`gas',`fluid',`solid',`mixed',etc), complexity (`simple',`complex'), normalcy (`normal',`abnormal'), caliber(`detailed',`normal'), malignancy_status (`benigh',`malignant'), patency (`patent',`occlusion'), occupancy (`empty'.`full'), integrity (`intact',`partially ruptured'). 

For example, the sentence `Similar 3.6 x 2.7 cm left thalamus dense hematoma with surrounding low-density vasogenic edema.' `3.6 x 2.7 cm' is `descriptor', `left thalamus' is `anatomy', `dense' is `descriptor', `hematoma' is `observation_present', `low-density' is `descriptor', `vasogenic edema' is `observation_present'. 
Note that for the sentence `Gray-white matter differentiation is within normal limits.', `Gray-white matter differentiation' should be  `observation_absent', and `within normal limits' should be `descriptor'. 
Note that for sentences include `loss of gray-white differentiation', should be `observation_present'. Note that for `CT',`MRI',`contrast administration',`post-contract',`clinical follow-up' are not procedure. for sentences that do not describe specific patient imaging findings but instead comment on the general capabilities or limitations of noncontrast CT, such as `Acute infarct may be inapparent on noncontrast CT.'. return an empty object {}. 
Note that observation should not contain the anatomy and descriptor.
Given a sentence, and reply with the JSON format following template: {`<sentence>':{`entity':`entity type',`entity':`entity type'}}
\end{lstlisting}

\begin{lstlisting}[style=customstyle]
You are a radiologist performing entity relation extraction from the FINDINGS and IMPRESSION sections in the given radiology report. Here a entity can be in [`anatomy',`observation_present',`observation_absent',`device_present',`device_absent',`procedure',`descriptor']. 
The relation can be in [`modify',`located_at',`suggestive_of',`associate_with']. 
The relation between entities can be  descriptor `modify' anatomy, descriptor `modify' observation, descriptor `modify' device observation/device `located_at' anatomy, observation/device `suggestive_of' observation/device, observation/device `associate_with' observation/device. 

For example, the sentence `Similar 3.6 x 2.7 cm left thalamus dense hematoma with surrounding low-density vasogenic edema.' `3.6 x 2.7 cm' is `descriptor', `left thalamus' is `anatomy', `dense' is `descriptor', `hematoma' is `observation_present'. `low-density' is `descriptor', `vasogenic edema' is `observation_present'. It should be `3.6 x 2.7 cm' modify `hematoma', `dense' modify `hematoma', `hematoma' located_at `left thalamus', `low-density' modify `vasogenic edema', `vasogenic edema' located_at `left thalamus'. 
    `Skull: No acute calvarial abnormality.', `Skull' is `anatomy', `No' is `descriptor', `acute' is `descriptor', `calvarial' is `anatomy', `abnormality' is `observation_absent'. It should be `No' modify `abnormality', `acute' modify `abnormality', `abnormality' located_at `calvarial'.

Given a radiology report, with entities and their types for each sentence like {`<sentence>':{`entity':`entity type',`entity':`entity type'},`<sentence>':{`entity':`entity type',`entity':`entity type'}  , determine the relationships between these entities. 

Reply with the extracted relationships in the following JSON format: {`<sentence>':[[source_entity,relation,target_entity],[...]],`<sentence>':[[source_entity,relation,target_entity],[...]]}
\end{lstlisting}

\end{document}